\documentclass[]{jingdong}

\usepackage{amsmath,amssymb,float}
\usepackage{array,tabularx}
\usepackage{url}
\usepackage{wrapfig}
\usepackage{amsthm}
\newtheorem{theorem}{Theorem}[section]

\microtypesetup{expansion=false}
\AddToHook{cmd/appendix/before}{\clearpage}

\newcommand{\method}{\textsc{H-MSAE}}

\title{\method{}: Hierarchical Sparse Concept Coverage for Efficient and Diverse Instruction Data Selection}

\author[*]{Bohan Cui}
\author[*,\dagger]{Jingyi Cui}
\author{Hongwei Wen}
\author{Feng Wang}
\author{Simiu Gu}
\affiliation{JD.com}
\contribution[*]{Equal contribution}
\contribution[\dagger]{Corresponding author}
\correspondence{Jingyi Cui (\email{cuijingyi.12@jd.com}, \email{rubycui0113@gmail.com})}
\abstract{
	Large instruction-tuning corpora contain substantial semantic redundancy, motivating the selection of compact subsets that preserve useful supervision while reducing training cost. Existing diversity-aware selectors often rely on generated semantic tags and explicit graph structures, introducing considerable annotation and preprocessing overhead. We introduce H-MSAE, a hierarchical sparse instruction selection framework that organizes multi-resolution concepts learned by a Matryoshka sparse autoencoder into a coarse-to-fine hierarchy. Quality-weighted, diminishing-return coverage with ancestor propagation promotes high-quality and non-redundant semantic coverage, while pool-relative length constraints regulate subset composition. We motivate this design through a local quality-diversity analysis of subset SFT utility and examine it using example-quality and representation-space diversity diagnostics. On Tulu 3 data pool, H-MSAE achieves the highest Final Avg among the evaluated 50K selectors across scales and initialization. The selected subset also exhibits the highest mean quality and effective rank among the evaluated methods. By avoiding full-pool generative tagging and supporting efficient sparse selection, H-MSAE reduces measured end-to-end runtime by approximately $5\times$ relative to MIG.
}

\begin{document}
\raggedbottom
\maketitle

\section{Introduction}

Large instruction-tuning corpora contain substantial semantic redundancy, making it desirable to select compact subsets that preserve useful supervision while reducing training cost. Effective selection must account for both example quality and semantic diversity: high-quality examples can provide reliable supervision, while non-redundant examples can expose the model to a broader range of training signals. Existing approaches address these properties through learned quality scores, generated semantic tags, taxonomies, or graph-based information gain~\citep{alpagasus2024,selectit2024,deita2024,instag2023,mig2025,tags2026}. However, diversity-aware selectors often rely on expensive full-pool annotation and explicit semantic structure construction. This motivates learning the semantic structure needed for selection directly from frozen conversation representations.

We introduce \method{}, a hierarchical sparse instruction selector built on multi-resolution concepts learned by a Matryoshka sparse autoencoder (MSAE). Cross-resolution co-activations establish soft hierarchical relations between sparse concepts, allowing an example to contribute to both fine-grained features and their broader ancestors. We then maximize a concave coverage objective that rewards underrepresented concepts while introducing diminishing returns for redundant examples. Unlike tag-based approaches, \method{} does not require generated semantic labels or an explicit graph over training examples.

Semantic novelty alone, however, does not guarantee useful supervision: rare or noisy examples can receive high coverage gains. We therefore incorporate example quality directly into the coverage objective, allowing quality to influence both immediate marginal gains and subsequent concept saturation. Since the adopted quality scorer exhibits a preference toward shorter examples, we additionally impose pool-relative length constraints to regulate subset composition. These design choices are motivated by our local SFT analysis, which identifies complementary contributions from supervision quality and directional diversity to expected subset utility. We examine this interpretation empirically through example-quality and representation-space diversity diagnostics.

\begin{wrapfigure}{r}{0.52\textwidth}
	\vspace{-2.2em}
	\centering
	\includegraphics[width=\linewidth]{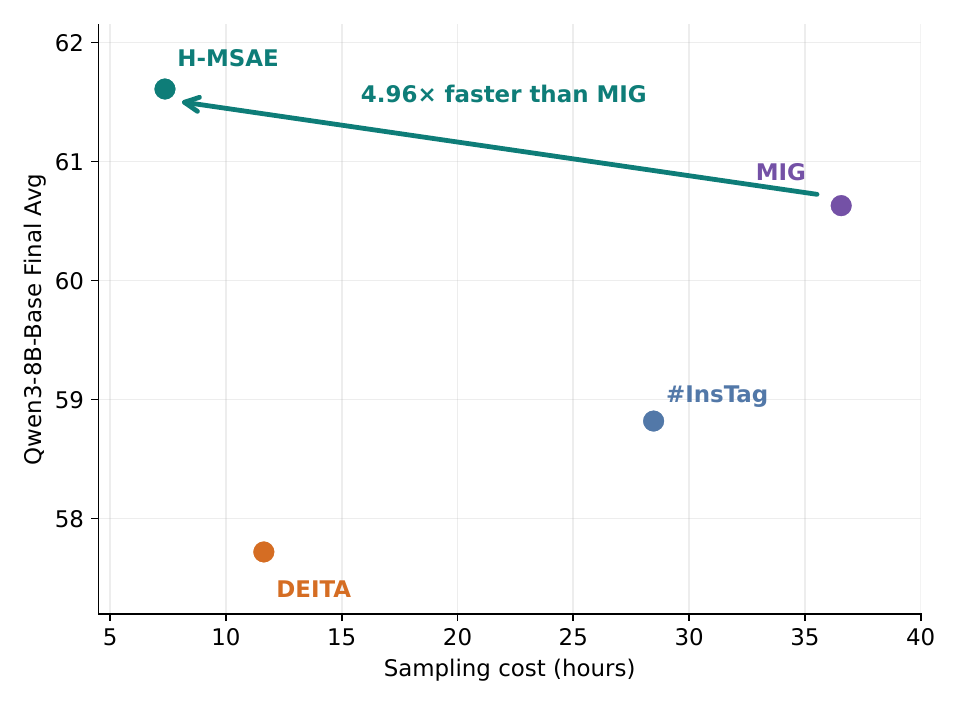}
	\caption{Performance--cost frontier on Qwen3-8B-Base. H-MSAE achieves a 0.98-point higher Final Avg than MIG while using approximately one-fifth of its end-to-end wall-clock time.}
	\label{fig:performance-efficiency}
	\vspace{-1.0em}
\end{wrapfigure}
Experiments on Tulu~3 show that \method{} achieves the highest Final Avg among the evaluated 50K selectors across three Qwen3 model scales, with particularly pronounced aggregate gains in subjective evaluation. Its selected subset also exhibits the highest mean quality and effective rank among the evaluated methods, while component ablations and data-scaling experiments support the combined design. Beyond selection effectiveness, \method{} achieves approximately $5\times$ lower end-to-end runtime than MIG by avoiding full-pool generative tagging. With prerequisite representations cached, lightweight hierarchy construction and sparse stochastic-greedy selection further reduce structure-construction and core sampling time by $4.8\times$ and $44.9\times$, respectively, consistent with the theoretical complexity analysis.

Our contributions are:
\begin{itemize}
    \item \textbf{Hierarchical sparse instruction selection with quality and composition control.}
    We introduce \method{}, which builds on multi-resolution sparse concepts obtained from a Matryoshka sparse autoencoder and organizes them into a coarse-to-fine hierarchy through cross-resolution relations. We perform diminishing-return coverage over this hierarchy with ancestor propagation, and further integrate quality-weighted coverage and pool-relative length constraints into a unified stochastic-greedy framework that jointly controls semantic coverage, example quality, and subset composition.

    \item \textbf{A quality--diversity view of instruction subset utility.}
    We interpret instruction subset utility through the joint role of example quality and semantic diversity: useful subsets should retain diverse semantic structure while avoiding low-quality supervision. This view is motivated by a local utility analysis and examined empirically through example quality and effective rank.

    \item \textbf{Strong downstream performance with efficient reusable selection.}
    \method{} achieves the highest overall average among the evaluated 50K selectors across model scales and initialization settings.
    The learned sparse representation enables lightweight post-representation selection and can be reused across selection budgets and downstream models. Unlike tag-based selectors, \method{} avoids full-pool generative tagging and explicit tag-level graph construction, substantially reducing the measured end-to-end selection cost.
\end{itemize}

\section{Related Work}

\paragraph{Instruction data selection.}
Instruction selectors use external judges, uncertainty, learned quality/complexity, open-set intents, gradients, or diversity objectives~\citep{alpagasus2024,selectit2024,deita2024,instag2023,less2024}.
DEITA combines quality and complexity scoring, while \#InsTag uses open-set semantic tags.
MIG combines tags and quality in a graph-based information-gain objective, and TAGS constructs a fine-grained taxonomy~\citep{mig2025,tags2026}.
At larger scale, sophisticated selectors do not always dominate random selection, making strong random controls important~\citep{ivison2025largescale}.
Our focus is a fixed-pool, downstream-label-free selector that replaces tag-based semantic structure with reusable sparse concepts.

\paragraph{Sparse autoencoders and representations.}
Sparse autoencoders (SAEs) learn sparse latent features from unlabeled neural representations, traditionally through sparsity regularization~\citep{ng2011sparse} and more recently through architectures such as $k$-sparse/Top-$k$ autoencoders~\citep{makhzani2013k,gao2024scaling}. Recent work has further studied the semantic quality, robustness, and theoretical limitations of SAE features~\citep{minegishi2025rethinking,cui2026limits}. SAE representations have also been explored for diversity-aware tuning and data engineering~\citep{saediversity2025,saerl2026}. Matryoshka Sparse Autoencoders (MSAEs) provide nested sparse representations at multiple capacities~\citep{bussmann2025learning}, which we adopt as the representation backbone of \method{}. On top of MSAE, we organize the resulting multi-resolution concepts through cross-resolution relations, optimize quality-weighted hierarchical concave coverage, and incorporate length-aware composition constraints for instruction selection.

\paragraph{Length and compute in instruction tuning.}
Response length is a nontrivial selection variable: selecting long responses can itself be a strong baseline~\citep{longismore2024}, and packing makes realized optimization steps depend on the selected length distribution.
Compute-aware data selection can also be posed explicitly~\citep{cads2025}.
Our length mechanism is narrower: it constrains subset composition using only pool statistics rather than optimizing against a target token total.

\section{\method{}: Hierarchical Sparse Instruction Selection}
\label{sec:method}

\begin{figure*}[t!]
\centering
\includegraphics[width=0.98\textwidth, trim=10 60 110 90, clip]{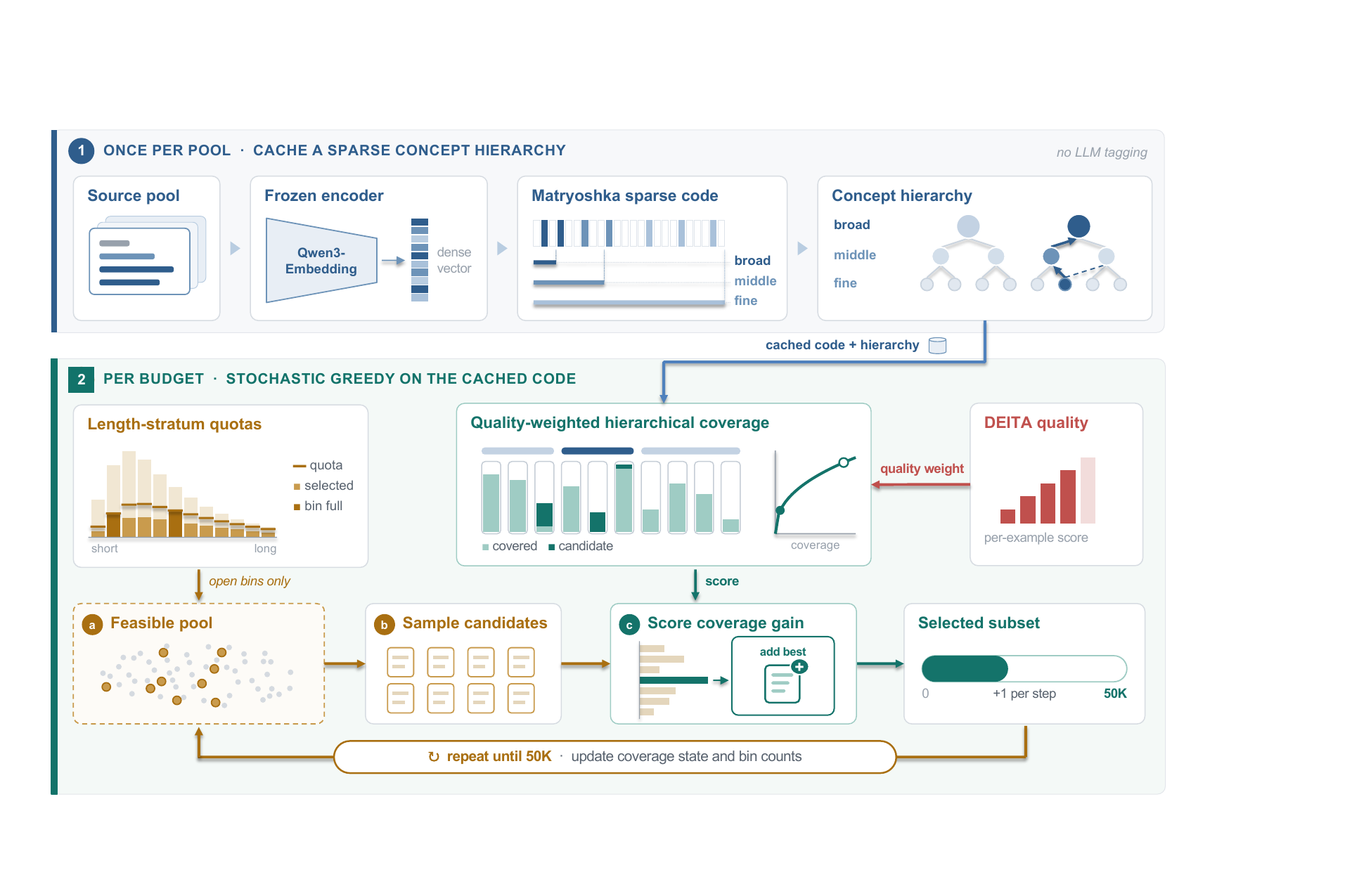}
\caption{Overview of \method{}. Frozen conversation embeddings are converted into multi-resolution sparse concepts and organized through cross-resolution relations. Quality-weighted hierarchical coverage determines semantic utility, while pool-derived length quotas constrain subset composition; stochastic greedy selects the final subset.}
\label{fig:pipeline}
\end{figure*}

\subsection{Problem Setup and Overview}

Let $\mathcal D=\{x_i\}_{i=1}^{N}$ be an eligible instruction-tuning pool and let $B\ll N$ denote the selection budget. We seek an index set $\mathcal I\subseteq[N]$ with $|\mathcal I|=B$ that preserves broad semantic coverage while avoiding redundant or low-quality supervision, without access to downstream evaluation labels.

\method{} consists of three components. First, we obtain multi-resolution sparse concepts using MSAE and organize them into a cross-resolution hierarchy. Second, we define a quality-weighted diminishing-return coverage objective over these concepts. Third, we maximize this utility using stochastic greedy under pool-derived length constraints.

\subsection{Hierarchical Sparse Concept Representation}
\label{sec:hierarchical-representation}

\paragraph{Multi-resolution sparse concepts.}
Each complete conversation is serialized and encoded using a frozen representation model as $e_i=E(\operatorname{serialize}(x_i))\in\mathbb R^d$. We adopt the Matryoshka Sparse Autoencoder (MSAE)~\citep{bussmann2025learning}, which learns nested sparse representations at multiple widths within a shared model. It exposes coarse-to-fine sparse concepts from the same representation space, providing a natural basis for constructing cross-resolution semantic relations.
A shared encoder produces the maximum-width sparse activation $z_i=\operatorname{BatchTopK}(\sigma(W_{\mathrm{enc}}e_i+b_{\mathrm{enc}}))$ with nested widths $D_1<\cdots<D_L$. The Matryoshka sparse autoencoder is trained by
\begin{equation}
\mathcal L_{\mathrm{MSAE}}
=
\frac{1}{N}
\sum_{i=1}^{N}
\sum_{\ell=1}^{L}
\alpha_\ell
\left\|
e_i
-
W_{\mathrm{dec}}^{(\ell)}
z_i^{(1:D_\ell)}
\right\|_2^2.
\label{eq:msae}
\end{equation}
Only newly introduced latent units are treated as concepts at each resolution: $\mathcal F_1=\{1,\ldots,D_1\}$ and $\mathcal F_\ell=\{D_{\ell-1}+1,\ldots,D_\ell\}$ for $\ell>1$. Thus, each latent concept belongs to exactly one resolution. 

\paragraph{Cross-resolution hierarchy.}
For concept $f$, let $\mathcal S_f=\{i:z_{if}>0\}$ denote its activation support. For a child concept $f\in\mathcal F_\ell$ and candidate parent $p\in\mathcal F_{\ell-1}$, we define asymmetric containment and the corresponding soft parent weight by
\begin{equation}
\operatorname{cont}(f,p)
=
\frac{|\mathcal S_f\cap\mathcal S_p|}
{|\mathcal S_f|},
\qquad
a_{fp}
=
\frac{\exp(\operatorname{cont}(f,p)/T_p)}
{\sum_{p'\in\mathcal P_{K_p}(f)}
\exp(\operatorname{cont}(f,p')/T_p)},
\label{eq:hierarchy}
\end{equation}
where $\mathcal P_{K_p}(f)$ contains the top-$K_p$ candidate parents in the immediately coarser resolution and $T_p$ is the parent-assignment temperature.

We propagate fine-grained activations toward their ancestors so that one example can contribute simultaneously to specific and broader concepts:
\begin{equation}
\widetilde z_{iu}
=
z_{iu}
+
\sum_{f\in\operatorname{Desc}(u)}
\pi_{f\rightarrow u}z_{if},
\label{eq:propagation}
\end{equation}
where $\operatorname{Desc}(u)$ denotes all direct and indirect descendants of concept $u$, and $\pi_{f\rightarrow u}$ is the total propagation weight along all paths from $f$ to $u$, determined by the soft parent weights and optional depth decay. The propagated activation $\widetilde z_i$ is used directly by the selection objective below.

\subsection{Quality-Weighted Hierarchical Coverage}
\label{sec:quality-coverage}

Let $q_i$ denote the quality score of example $i$ and define its normalized quality weight as $\omega_i=(q_i/\bar q)^\beta$, where $\bar q=N^{-1}\sum_{i=1}^{N}q_i$ and $\beta\ge0$ controls the strength of quality weighting. We use the released DEITA quality scorer in our implementation \citep{deita2024}.

The subset utility is defined by quality-weighted concave coverage across all hierarchy levels:
\begin{equation}
F(\mathcal I)
=
\sum_{\ell=1}^{L}
\lambda_\ell
\sum_{f\in\mathcal F_\ell}
\phi\!\left(
\sum_{i\in\mathcal I}
\omega_i\widetilde z_{if}
\right),
\qquad
\phi(t)
=
(t+\epsilon)^\gamma-\epsilon^\gamma,
\quad
0<\gamma<1.
\label{eq:coverage}
\end{equation}
Here $\widetilde z_{if}$ determines which concepts an example covers across resolutions, while $\omega_i$ controls the strength of its contribution. The concavity of $\phi$ introduces diminishing returns, favoring examples that cover insufficiently represented concepts over repeated coverage of already saturated regions. We denote the marginal gain of candidate $i$ by 
\begin{equation}
\Delta_F(i\mid \mathcal I)=F(\mathcal I\cup\{i\})-F(\mathcal I).    
\end{equation}

\subsection{Length-Constrained Stochastic-Greedy Selection}
\label{sec:stochastic-greedy}

Because quality-aware selection can alter the response-length distribution, particularly when the DEITA quality scorer exhibits a preference toward shorter examples, we use length to constrain subset composition rather than incorporating it into the semantic utility. We partition the eligible pool into $m$ quantile bins $\{\mathcal B_b\}_{b=1}^{m}$ according to serialized conversation length and define the feasible family
\begin{equation}
\mathcal M
=
\left\{
\mathcal I\subseteq[N]:
|\mathcal I\cap\mathcal B_b|
\le\kappa_b,\ \forall b
\right\},
\qquad
\kappa_b\propto\bar L_b^{\,\tau},
\qquad
\sum_{b=1}^{m}\kappa_b=B,
\label{eq:length-constraint}
\end{equation}
where $\bar L_b$ is the mean serialized length in bin $b$ and $\tau\ge0$ controls the strength of length reweighting. 

Starting from $\mathcal I_0=\varnothing$, at iteration $t$ we sample a candidate set $R_t$ from the currently feasible examples $\{i\notin\mathcal I_t:\mathcal I_t\cup\{i\}\in\mathcal M\}$ and select
\begin{equation}
i_t^\star
=
\arg\max_{i\in R_t}
\Delta_F(i\mid \mathcal I_t),
\qquad
\mathcal I_{t+1}
=
\mathcal I_t\cup\{i_t^\star\}.
\label{eq:selection}
\end{equation}
Following \citet{mirzasoleiman2015}, we use at most $s=\lceil(N/B)\log(1/\varepsilon_{\mathrm{sg}})\rceil$ sampled candidates per iteration. The standard approximation guarantee associated with this rule applies to the cardinality-constrained setting; under our length-aware constraints, we use the same sampling rule with explicit feasibility checks.
After $B$ iterations, the algorithm returns the final selected index subset $\mathcal I_B$.

\section{Analysis of Subset Utility and Efficiency}

\subsection{A Quality--Diversity Analysis of Subset SFT}
\label{sec:subset-utility}

We analyze why \method{} can yield effective instruction subsets from a quality--diversity perspective. Its objective combines quality weighting with diminishing-return coverage over hierarchical sparse concepts, encouraging both reliable supervision and non-redundant semantic content. We show that, under a local SFT model, these two properties contribute complementary positive terms to expected target-risk reduction.

Let $\theta_0$ denote the model before SFT and
\begin{equation}
R_{\mathcal T}(\theta)
=
\mathbb E_{z\sim\mathcal T}[\ell(z;\theta)]
\end{equation}
the population risk on a target instruction distribution $\mathcal T$. We denote its local gradient and Hessian by
$g_{\mathcal T}
=
\nabla_\theta R_{\mathcal T}(\theta_0)$, and
$
H_{\mathcal T}
=
\nabla_\theta^2 R_{\mathcal T}(\theta_0)
\succeq 0$.
For a fixed-size subset $S$, $|S|=B$, let $g_i$ denote the local per-example training gradient, modeled stochastically below. We approximate the update induced by subset SFT as
$\Delta_S
=
-\frac{\eta}{B}\sum_{i\in S} g_i$,
where $\eta>0$ is an effective local step size, and define the utility of $S$ by its reduction in target risk,
\begin{equation}
U_{\mathcal T}(S)
=
R_{\mathcal T}(\theta_0)
-
R_{\mathcal T}(\theta_0+\Delta_S).
\label{eq:subset-utility-qdiv}
\end{equation}

\paragraph{Local gradient model.}
The population gradient $g_{\mathcal T}$ specifies the local direction of the target objective, whereas an individual training example contributes a stochastic gradient $g_i$ that need not coincide with $g_{\mathcal T}$. We write $\bar g_i:=\mathbb E[g_i]$ for its expected training direction and associate each example with a latent quality $q_i\in[0,1]$, where higher quality corresponds to lower stochastic uncertainty around $\bar g_i$. The precise covariance model are given in Appendix~\ref{app:quality-diversity-proof}.

\paragraph{Subset quality.}
For a subset $S$ of size $B$, we define its average quality as
\begin{equation}
Q(S)
=
\frac{1}{B}\sum_{i\in S}q_i.
\label{eq:subset-quality}
\end{equation}
$Q(S)$ measures the reliability of the supervision contained in the subset: a larger value means that, on average, the selected examples induce training gradients with lower stochastic uncertainty.

\paragraph{Directional diversity.}
To characterize redundancy between examples, we compare their expected training directions in the local geometry induced by the target-risk Hessian $H_{\mathcal T}$. For each example, define the curvature-normalized direction as $v_i = H_{\mathcal T}^{1/2}\bar g_i / \|H_{\mathcal T}^{1/2}\bar g_i\|_2$. We define the directional diversity of a subset $S$ as
\begin{equation}
D(S)
=
\operatorname{tr}\!\left(
\widehat{\operatorname{Cov}}_{i\in S}[v_i]
\right),
\label{eq:subset-diversity}
\end{equation}
where $\operatorname{tr}(\cdot)$ denotes the trace and $\widehat{\operatorname{Cov}}_{i\in S}[v_i]$ denotes the sample covariance of the normalized directions $\{v_i\}_{i\in S}$. 
Intuitively, $D(S)$ measures their total directional dispersion: it is small when the selected examples induce highly redundant training directions and increases as these directions spread across the local target-risk geometry.

Throughout the analysis, $\mathbb E_g$ denotes expectation over the stochastic training gradients $\{g_i\}_{i\in S}$ for a fixed subset $S$. Thus, $Q(S)$ and $D(S)$ are deterministic properties of the selected examples, while their contributions to expected utility can vary with the choice of $S$.

\begin{theorem}[Quality--diversity decomposition of subset SFT utility]
\label{thm:quality-diversity}
Under the local gradient modeling, the expected target-risk reduction of a given subset $S$ of size $B$ admits the decomposition
\begin{equation}
\mathbb E_g[U_{\mathcal T}(S)]
=
C_B
+
\lambda_Q Q(S)
+
\lambda_D D(S)
+
o(\eta^2)
\label{eq:quality-diversity-utility}
\end{equation}
where $C_B$, $\lambda_Q$, $\lambda_D$ are specified in the proof.
\end{theorem}

Theorem~\ref{thm:quality-diversity} shows that expected subset utility increases with both average gradient reliability and directional dispersion. Here, $\mathbb E_g$ denotes expectation over the stochastic training gradients $\{g_i\}_{i\in S}$ for a fixed subset $S$. Thus, $Q(S)$ and $D(S)$ are deterministic properties of the selected examples and can vary with the subset composition. The proof is provided in Appendix~\ref{app:quality-diversity-proof}.

This decomposition provides an interpretation of the two signals combined by \method{}: quality-weighted coverage favors examples with higher estimated supervision quality, while diminishing-return hierarchical coverage promotes non-redundant coverage across latent semantic regions. 
{
This decomposition relies on a local second-order approximation of SFT and assumptions on gradient alignment, self-curvature, and quality-dependent gradient covariance. It provides theoretical motivation for the quality--diversity trade-off rather than a direct guarantee for the proposed hierarchical coverage objective, since the practical quality and semantic coverage signals are not formally established to approximate the gradient-based quantities in the theorem. 
Empirically,} since the implementation uses DEITA quality scores, we additionally impose length constraints to mitigate their bias toward shorter examples. 
In Section~\ref{sec:quality-diversity-diagnostics}, we empirically examine whether the selected subsets exhibit high example quality and broad representation-space diversity.

\subsection{Computational Complexity and Efficiency}
\label{sec:complexity}

We analyze the efficiency of \method{} from three complementary aspects: lightweight hierarchy construction, sparse subset selection, and the avoidance of full-pool generative annotation.

\paragraph{Lightweight hierarchy construction.}
\method{} constructs cross-resolution relations by accumulating sparse co-activations between adjacent resolutions and assigning each child concept to a bounded constant number of candidate parents. With at most $O(k)$ active concepts per example, each example contributes at most $O(k^2)$ active cross-resolution pairs. Over a pool of $N$ examples, hierarchy construction therefore requires approximately
\begin{equation}
O(Nk^2+E),
\end{equation}
where $E$ is the number of distinct observed cross-resolution concept pairs. Since each non-root concept has a constant number of parents, the final hierarchy contains $O(D)$ edges and requires $O(D)$ storage, with $O(E)$ temporary storage during construction. In contrast to MIG's explicit graph over generated semantic tags, \method{} directly constructs a compact hierarchy over learned sparse concepts.

\paragraph{Sparse subset selection.}
Given the hierarchy, marginal-gain computation only accesses the active concepts of a candidate and their propagated ancestors. With $k\ll D$ active concepts and $L$ hierarchy resolutions, the per-candidate cost is approximately $O(kL)$. Stochastic greedy evaluates only
$
s=\left\lceil\frac{N}{B}\log\frac{1}{\varepsilon_{\mathrm{sg}}}\right\rceil
$
candidates at each of the $B$ selection steps, yielding an approximate selection cost of
\begin{equation}
O(BskL).
\end{equation}
Length-aware quotas add only constant-time feasibility checks when bin assignments and remaining capacities are maintained. The sampling budget follows standard stochastic greedy; we do not assume that its original cardinality-constrained approximation guarantee directly extends to the length-constrained setting.

\paragraph{Avoiding full-pool annotation.}
The above analysis concerns hierarchy construction and subset selection. End-to-end cost also depends on how the semantic representation is obtained. Tag-based selectors require generative annotation over the candidate pool, whereas \method{} learns sparse concepts from frozen conversation embeddings through an MSAE and therefore avoids full-pool generative tagging. This provides an additional source of end-to-end efficiency beyond sparse hierarchy construction and selection.

Section~\ref{sec:efficiency} empirically evaluates these advantages through end-to-end subset-construction time.

\section{Experiments}
\label{sec:experiments}

\subsection{Experimental Setup}

We select 50K examples from the 939K-example Tulu~3 pool and compare
\method{} with Uniform Random \citep{xia2025rethinking}, ZIP \citep{yin2024entropy}, IFD \citep{li2024quantity}, \#InsTag \citep{instag2023}, DEITA \citep{deita2024}, CaR \citep{ge2024clustering}, QDIT \citep{bukharin2024data},
and MIG \citep{mig2025}. Full Tulu~3 is included only as a full-data reference.

Our primary comparisons use three-epoch full-parameter SFT with a 4K
context window and packing. We evaluate six objective benchmarks and
three subjective benchmarks; Avg$_{\rm obj}$ and Avg$_{\rm sub}$ denote
their respective aggregates, and Final Avg is their equally weighted mean.
Detailed model, training, selection, and evaluation configurations are
provided in Appendix~\ref{app:experimental-settings}.

\subsection{Main Results}
\label{sec:main-results}

\begin{table*}[h]
\centering
\caption{Qwen3-Base results across 1.7B, 4B, and 8B scales. All rows use 50K selected examples unless marked Full. Best, second-best distinct, and third-best distinct comparable 50K results are shown in bold, underline, and italics, respectively.}
\vspace{-2mm}
\label{tab:qwen-base-main}
\label{tab:strict-main}
\begingroup
\renewcommand{\arraystretch}{0.42}
\resizebox{\textwidth}{!}{%
\begin{tabular}{lllrrrrrrr|rrrrr}
\toprule
Base Model & Method & Data Size & ARC & BBH & GSM & HE & MMLU & IFEval & Avg$_{\rm obj}$ & AE & MT & Wild & Avg$_{\rm sub}$ & Avg \\
\midrule
\multirow{10}{*}{Qwen3-1.7B-Base}
 & Full Tulu~3 & 939K & 76.27 & 55.29 & 66.57 & 61.59 & 61.66 & 56.75 & 63.02 & 2.36 & 5.70 & -60.71 & 26.33 & 44.68 \\
\cmidrule(l){2-15}
 & Uniform Random & 50K & 73.56 & 57.41 & 69.37 & 57.93 & 59.11 & \underline{51.76} & 61.52 & 1.61 & 5.66 & -65.64 & 25.13 & 43.33 \\
 & ZIP & 50K & 74.24 & 57.26 & 69.14 & \underline{60.98} & 59.73 & 47.32 & 61.45 & 1.24 & 5.84 & -71.71 & 24.59 & 43.02 \\
 & IFD & 50K & 74.24 & 58.11 & 67.55 & 59.76 & 61.42 & 48.24 & 61.55 & 2.73 & 5.74 & \textbf{-57.47} & 27.13 & 44.34 \\
 & \#InsTag & 50K & \textit{74.58} & 57.73 & 66.64 & 56.10 & 60.02 & 49.72 & 60.80 & 2.11 & 5.99 & -63.17 & 26.81 & 43.80 \\
 & DEITA & 50K & 72.88 & 55.69 & 69.83 & \textbf{61.59} & 59.70 & 48.80 & 61.41 & 1.49 & \textit{6.14} & -65.04 & 26.79 & 44.10 \\
 & CaR & 50K & \textbf{77.63} & \textbf{58.34} & 68.54 & 56.71 & 61.44 & 49.91 & \underline{62.10} & \underline{3.11} & 5.71 & -58.60 & 26.97 & 44.53 \\
 & QDIT & 50K & \underline{76.27} & 57.63 & \textit{70.36} & 57.32 & \underline{61.66} & 48.43 & 61.95 & 2.61 & 6.06 & \textit{-57.83} & \textit{28.10} & \textit{45.02} \\
 & MIG & 50K & 72.20 & \underline{58.32} & \underline{71.49} & 57.32 & \textit{61.49} & \textit{51.02} & \textit{61.97} & \textit{2.98} & \textbf{6.39} & -61.78 & \underline{28.66} & \underline{45.32} \\
\cmidrule(l){2-15}
 & \textbf{\method{}} & 50K & 73.22 & \textit{58.13} & \textbf{72.25} & \textit{60.37} & \textbf{62.30} & \textbf{52.87} & \textbf{63.19} & \textbf{3.35} & \underline{6.38} & \underline{-57.78} & \textbf{29.42} & \textbf{46.31} \\
\midrule
\multirow{10}{*}{Qwen3-4B-Base}
 & Full Tulu~3 & 939K & 88.47 & 72.89 & 82.87 & 78.66 & 72.46 & 72.46 & 77.97 & 4.60 & 7.10 & -33.27 & 36.32 & 57.15 \\
\cmidrule(l){2-15}
 & Uniform Random & 50K & \textbf{86.44} & 75.54 & 83.47 & 68.90 & 71.68 & 72.09 & 76.35 & 3.85 & 7.10 & -41.31 & 34.73 & 55.54 \\
 & ZIP & 50K & 85.42 & 76.81 & 85.52 & \underline{73.17} & 71.04 & 71.16 & \textit{77.19} & 3.48 & 6.54 & -50.10 & 31.28 & 54.23 \\
 & IFD & 50K & \underline{86.10} & \textit{76.82} & 82.18 & \textbf{73.78} & \underline{73.20} & 68.58 & 76.78 & \underline{6.34} & 6.96 & -33.15 & 36.45 & 56.62 \\
 & \#InsTag & 50K & \textbf{86.44} & 75.89 & 84.61 & 69.51 & 71.73 & \textit{73.38} & 76.93 & \textit{6.21} & \underline{7.30} & -35.24 & \textit{37.20} & \textit{57.06} \\
 & DEITA & 50K & \textit{85.76} & 75.60 & \textbf{87.64} & 69.51 & 71.53 & 71.90 & 76.99 & 4.22 & 6.62 & -40.61 & 33.37 & 55.18 \\
 & CaR & 50K & \textit{85.76} & 75.12 & \textit{85.82} & 71.34 & 72.12 & 70.98 & 76.86 & 4.84 & 7.01 & \underline{-32.39} & 36.25 & 56.55 \\
 & QDIT & 50K & \textit{85.76} & \textbf{77.16} & 84.31 & 71.34 & \textit{73.08} & 71.53 & \underline{77.20} & 5.34 & 7.04 & \textit{-32.59} & 36.48 & 56.84 \\
 & MIG & 50K & 81.02 & \underline{76.87} & \underline{86.43} & 68.29 & 72.75 & \textbf{74.49} & 76.64 & \textbf{7.33} & \textbf{7.38} & -33.41 & \underline{38.14} & \underline{57.39} \\
\cmidrule(l){2-15}
 & \textbf{\method{}} & 50K & 82.71 & \textbf{77.16} & 83.62 & \textit{72.56} & \textbf{73.76} & \underline{74.31} & \textbf{77.35} & \textbf{7.33} & \textit{7.22} & \textbf{-25.07} & \textbf{39.00} & \textbf{58.18} \\
\midrule
\multirow{10}{*}{Qwen3-8B-Base}
 & Full Tulu~3 & 939K & 90.85 & 73.98 & 83.24 & 83.54 & 77.94 & 70.61 & 80.03 & 6.83 & 5.87 & -33.04 & 33.00 & 56.51 \\
\cmidrule(l){2-15}
 & Uniform Random & 50K & \textit{91.53} & 74.94 & 70.43 & \underline{85.98} & 66.74 & 64.88 & 75.75 & 5.59 & 7.41 & -25.93 & 38.91 & 57.33 \\
 & ZIP & 50K & 90.85 & 47.97 & 80.14 & 78.05 & \underline{78.18} & 48.61 & 70.63 & 1.74 & 6.58 & -43.22 & 31.98 & 51.30 \\
 & IFD & 50K & \textit{91.53} & \textbf{79.63} & \textbf{87.19} & 84.15 & 77.76 & 59.15 & 79.90 & \textit{10.81} & 7.29 & -24.64 & \textit{40.46} & 60.18 \\
 & \#InsTag & 50K & 91.19 & 73.66 & 82.18 & 83.54 & 77.99 & 64.70 & 78.88 & 5.22 & 7.31 & \textit{-24.05} & 38.76 & 58.82 \\
 & DEITA & 50K & 90.51 & 73.22 & 82.56 & 81.10 & \textbf{78.28} & \underline{66.91} & 78.76 & 5.09 & 7.19 & -33.88 & 36.68 & 57.72 \\
 & CaR & 50K & \textit{91.53} & 75.52 & 82.49 & 82.32 & 77.89 & 58.78 & 78.09 & 9.81 & \underline{7.51} & -28.86 & 40.16 & 59.12 \\
 & QDIT & 50K & 90.85 & \underline{78.87} & \underline{87.04} & \textbf{86.59} & 77.66 & 59.15 & \textit{80.03} & \underline{11.55} & \textit{7.42} & -27.00 & \underline{40.75} & \textit{60.39} \\
 & MIG & 50K & \underline{92.88} & \textit{78.49} & \textit{86.20} & \textit{85.37} & 77.53 & \textbf{67.84} & \textbf{81.38} & 8.57 & 7.25 & \underline{-22.87} & 39.88 & \underline{60.63} \\
\cmidrule(l){2-15}
 & \textbf{\method{}} & 50K & \textbf{93.22} & 78.44 & 84.08 & 81.10 & \textit{78.01} & \textit{66.73} & \underline{80.26} & \textbf{12.42} & \textbf{7.55} & \textbf{-18.10} & \textbf{42.96} & \textbf{61.61} \\
\bottomrule
\end{tabular}%
}
\vspace{-3mm}
\endgroup
\end{table*}

\paragraph{Consistent performance across scales.}
Table~\ref{tab:qwen-base-main} shows that \method{} achieves the highest Final Avg among the evaluated 50K selectors across all three Qwen3 scales, indicating that its effectiveness is not confined to a particular model size. Compared with competing selection methods, \method{} exhibits more pronounced gains in subjective evaluation than in objective performance across the three scales. This suggests that quality-weighted hierarchical coverage may be particularly beneficial for subjective evaluation, highlighting the importance of considering both dimensions when assessing instruction data selection.

\paragraph{Random and full-data controls.}
Across all three scales, \method{} consistently outperforms both Uniform Random and Full Tulu~3, despite using substantially fewer examples than the latter. These results highlight the importance of data selection beyond simply increasing training data volume under the reported training protocol. Additional results on different model initializations and architectures are provided in Appendix~\ref{app:additional-models}.

{
\textbf{Repeated-run evaluation.} In addition to the single-run benchmark results in Table \ref{tab:qwen-base-main}, we conduct three-seed experiments to assess the robustness of H-MSAE across model scales and architectures, shown in Table \ref{tab:hmsae_three_seed} in Appendix \ref{app:additional-models}. The additional results are reported as mean $\pm$ standard deviation over three independent training seeds. H-MSAE achieves positive mean Final Avg improvements over MIG on Qwen3-1.7B/4B/8B and Mistral-7B-v0.3, with gains of +1.05/+1.56/+0.62/+0.87 points, respectively.
}

\subsection{Selection Efficiency}
\label{sec:efficiency}

We measure the cost of selecting 50K examples from Tulu~3 using complete end-to-end pipelines and separately profile structure construction and core sampling after method-specific prerequisites are available. All experiments are conducted on a single B200 GPU with 32 CPU cores and 256\,GB of host memory.

\begin{table*}[h]
\centering
\small
\setlength{\tabcolsep}{5pt}
\caption{End-to-end and selection-stage efficiency for selecting 50K examples from the full Tulu~3 pool. Relative costs are normalized by \method{} within each timing scope. End-to-end measurements include all method-specific prerequisites. Running times are medians over three profiling runs.}

\label{tab:efficiency}
\resizebox{\linewidth}{!}{
\begin{tabular}{lrrlllrrrr}
\toprule
& \multicolumn{2}{c}{Accounted total}
& \multicolumn{3}{c}{Major prerequisite costs}
& \multicolumn{2}{c}{Structure construction}
& \multicolumn{2}{c}{Core sampling / filtering} \\
\cmidrule(lr){2-3}\cmidrule(lr){4-6}\cmidrule(lr){7-8}\cmidrule(lr){9-10}
Method & Total & Rel. & Embedding & LLM tagging & Scoring & Time & Rel. & Time & Rel. \\
\midrule
\method{}
& \textbf{7h22m} & 1.00$\times$
& 3h13m & -- & 4h02m
& \textbf{7.86 s} & 1.00$\times$
& 28.89 s & 1.00$\times$ \\
MIG
& 36h34m & 4.96$\times$
& -- & 28h28m & 7h40m
& 37.44 s & 4.76$\times$
& 1,296.16 s & 44.87$\times$ \\
\#InsTag
& 28h28m & 3.86$\times$
& -- & 28h28m & --
& -- & --
& \textbf{7.65 s} & 0.26$\times$ \\
DEITA
& 11h38m & 1.58$\times$
& 3h47m & -- & 7h40m
& -- & --
& 76.02 s & 2.63$\times$ \\
\bottomrule
\end{tabular}
}

\vspace{1mm}
\parbox{\linewidth}{\footnotesize
$^\dagger$In the prerequisite stage, H-MSAE requires additionally only 5m18s for MSAE training and sparse encoding. 
}
\vspace{-5mm}
\end{table*}

\paragraph{End-to-end efficiency.}
As shown in Table~\ref{tab:efficiency}, \method{} reduces end-to-end runtime by approximately $4.96\times$ relative to MIG and $3.86\times$ relative to \#InsTag. The main difference lies in prerequisite computation: \#InsTag and MIG require expensive full-pool generative tagging, while DEITA incurs substantial embedding and scoring costs. Although quality scoring and embedding account for approximately 98\% of \method{}'s own runtime, MSAE training and sparse encoding add only modest overhead. By learning sparse concepts from frozen conversation embeddings, \method{} avoids the full-pool generative annotation required by tag-based selectors. Moreover, \method{} uses only quality scores, whereas DEITA and MIG require both quality and complexity scores, reducing its scoring cost by approximately half. Notably, even without quality scoring, \method{} achieves a Final Avg of 60.54, comparable to MIG's 60.63, while reducing end-to-end runtime to approximately 3h20m, an $11\times$ speedup over MIG.

\paragraph{Structure construction and sampling.}
With prerequisite representations and annotations cached, \method{} further reduces structure-construction and core sampling time relative to MIG by $4.8\times$ and $44.9\times$, respectively. The former reflects lightweight cross-resolution relations derived from sparse co-activations, avoiding MIG's explicit tag-graph construction. The latter is consistent with the combination of stochastic greedy candidate evaluation and sparse marginal-gain computation, which accesses only active concepts and their propagated ancestors. These results support the two sources of computational savings discussed in Section~\ref{sec:complexity}: avoiding expensive full-pool annotation and reducing the cost of structure construction and subset selection through sparse computation. While \#InsTag employs a faster core sampling procedure, its overall runtime is dominated by generative tagging.

\subsection{Quality and Diversity Diagnostics}
\label{sec:quality-diversity-diagnostics}

Our local utility analysis in Section~\ref{sec:subset-utility} suggests that example quality and directional diversity contribute complementary terms to expected subset utility. To examine whether the selected subsets exhibit these desirable properties, we use the official DEITA quality score as a proxy for supervision quality and the effective rank (eRank) of frozen conversation embeddings as a measure of representation-space diversity. These empirical metrics are motivated by, but do not directly measure, the gradient-based quantities $Q(S)$ and $D(S)$ in Theorem~\ref{thm:quality-diversity}.

\begin{figure*}[h]
	\centering
	
	\begin{minipage}[t]{0.43\textwidth}
		\centering
		\captionsetup{type=table}
		\captionof{table}{
			Quality and diversity diagnostics on the same frozen 50K selections.
			Q mean is the average DEITA quality score,
			while eRank measures the effective rank of the selected representation space.
		}
		\label{tab:subset-priority-metrics}
		   \vspace{-2mm}
		
		\small
		\setlength{\tabcolsep}{8pt}
		\begin{tabular}{lrr}
			\toprule
			Method & Q mean $\uparrow$ & eRank $\uparrow$ \\
			\midrule
			Random & 3.848 & 156.07 \\
			\#InsTag & 4.049 & 154.10 \\
			DEITA & 3.776 & 150.12 \\
			MIG & 4.598 & 149.13 \\
			H-MSAE & \textbf{4.993} & \textbf{158.15} \\
			\bottomrule
		\end{tabular}
        \vspace{-3mm}
	\end{minipage}
	\hfill
	\begin{minipage}[t]{0.54\textwidth}
		\centering
		\captionsetup{type=figure}
		\captionof{figure}{
			Data scaling on Qwen3-1.7B-Base.
			Final Avg across selection budgets of 10K, 20K, 50K, and 100K,
			with Full Tulu~3 as a reference.
		}
		\label{fig:qwen-scaling-main}
		   \vspace{-2mm}
		
		\includegraphics[
		width=\linewidth,
		trim=0 0 0 0,
		clip
		]{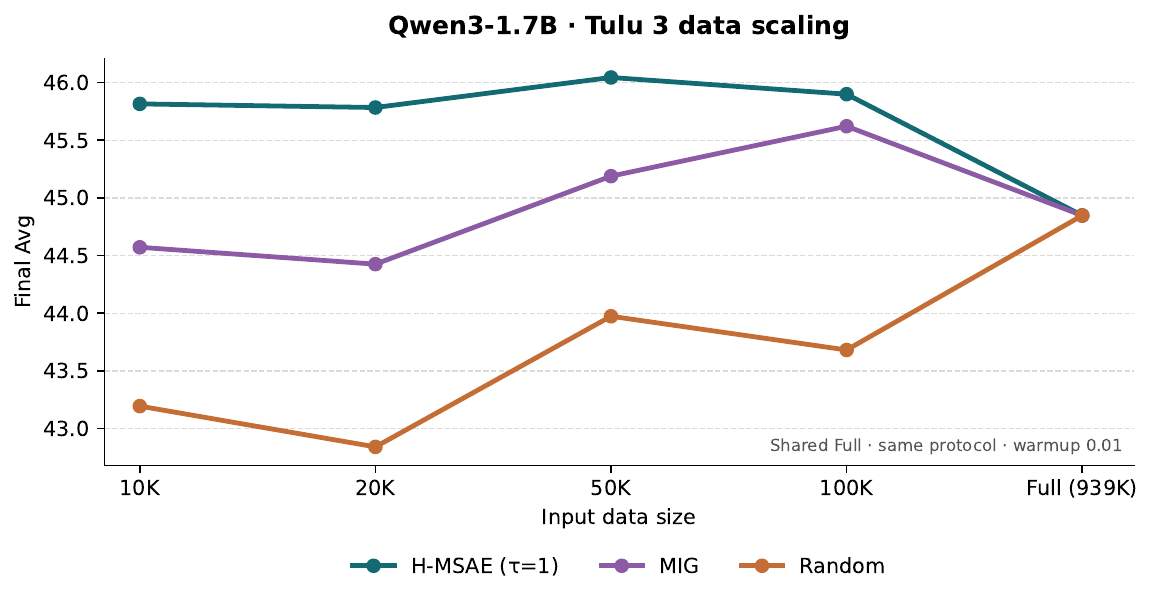}
	    \vspace{-10mm}
    \end{minipage}
	
\end{figure*}

Table~\ref{tab:subset-priority-metrics} reveals distinct quality--diversity profiles among the evaluated selectors. Random selection retains relatively broad representation-space diversity but has lower average quality, whereas MIG selects higher-quality examples with more concentrated representations. In contrast, \method{} achieves the highest mean quality and eRank simultaneously, indicating that quality-weighted hierarchical coverage can favor high-quality examples without sacrificing broad latent support.

These findings are consistent with the quality--diversity interpretation in Section~\ref{sec:subset-utility}, providing empirical support for combining quality weighting with diminishing-return hierarchical coverage.

\subsection{Ablation and Scaling}
\label{sec:component-ablation}

\paragraph{Component ablation.}
We conduct controlled ablations on Qwen3-1.7B to examine four components of \method{}: multi-resolution reconstruction, ancestor propagation, quality weighting, and length control. All variants use the same frozen conversation embeddings, candidate pool, 50K selection budget, and 256-step SFT protocol. We remove quality weighting and length control individually and jointly, disable ancestor propagation, and separately ablate multi-resolution reconstruction while retaining the MSAE architecture.

Table~\ref{tab:component-ablation} shows that the full configuration achieves the highest overall Avg. Removing either quality weighting or length control reduces performance, with their joint removal producing the largest degradation. This is consistent with their complementary roles: quality weighting favors informative supervision, while length constraints regulate the resulting subset composition. Disabling ancestor propagation also reduces both evaluation aggregates, supporting the use of cross-resolution relations rather than treating sparse concepts independently. Finally, multi-resolution reconstruction improves the overall result primarily through objective evaluation, while its removal yields a slightly higher subjective score. These findings support the combined design, although the contributions of individual components differ across evaluation dimensions.

\begin{table*}[h]
\centering
\small
\caption{Controlled component ablation on Qwen3-1.7B.
The multi-resolution reconstruction ablation removes the broad- and middle-resolution reconstruction objectives by assigning reconstruction loss weights $(0,0,1)$.}
\label{tab:component-ablation}
\vspace{-2mm}
\resizebox{\linewidth}{!}{
\begin{tabular}{lccccccc}
\toprule
Variant
& Multi-res. recon.
& Ancestor prop.
& Quality
& Length
& Avg$_{\mathrm{obj}}$ $\uparrow$
& Avg$_{\mathrm{sub}}$ $\uparrow$
& Avg $\uparrow$ \\
\midrule
\textbf{Full \method{}}
& \checkmark & \checkmark 
& \checkmark & \checkmark
& \textbf{63.53} & 30.06 & \textbf{46.79} \\

w/o quality
& \checkmark & \checkmark 
& $\times$ & \checkmark
& 62.60 & 28.11 & 45.35 \\

w/o length control
& \checkmark & \checkmark 
& \checkmark & $\times$
& 62.56 & 28.84 & 45.70 \\

w/o quality \& length
& \checkmark & \checkmark 
& $\times$ & $\times$
& 62.71 & 27.44 & 45.07 \\

w/o multi-res. recon.
& $\times$ & \checkmark 
& \checkmark & \checkmark
& 62.36 & 30.23 & 46.29 \\

w/o ancestor prop.
& \checkmark & $\times$ 
& \checkmark & \checkmark
& 62.56 & 29.27 & 45.92 \\

\bottomrule
\end{tabular}
}
\vspace{-3mm}
\end{table*}

\paragraph{Data scaling.}
We independently rerun \method{}, MIG, and Uniform Random at selection budgets of 10K, 20K, 50K, and 100K, recomputing the length quotas and selected subsets for each budget. As shown in Figure~\ref{fig:qwen-scaling-main}, \method{} consistently outperforms both baselines across the evaluated budgets, indicating that its effectiveness is not specific to the default 50K setting. Performance improves substantially from 10K to 50K examples but largely saturates thereafter, suggesting diminishing returns from further increasing the selection budget under the current training protocol.

\section{Conclusion}

We introduced \method{}, a hierarchical sparse instruction selection framework that combines multi-resolution semantic coverage, quality weighting, and length-aware subset composition. On Tulu~3, \method{} achieves the highest Final Avg among the evaluated 50K selectors across three scales and an instruction-tuned initialization. Quality and diversity diagnostics, component ablations, and data-scaling experiments further support the combined design. In addition to its downstream performance, \method{} reduces end-to-end selection time by approximately $5\times$ relative to MIG, with substantially lower structure-construction and core sampling costs. Together, these results demonstrate the potential of hierarchical sparse concepts as an effective and computationally efficient alternative to generated semantic tags and explicit tag-graph construction for instruction data selection. Future work will investigate the transferability of learned sparse concepts across instruction domains and model architectures, as well as their application to larger-scale and dynamically evolving training corpora.

\section*{AI Use Statement}

In this work, we used generative AI tools to assist with intermediate mathematical derivations and the writing and checking of parts of the theoretical proofs, the implementation and debugging of the proposed method and experimental code, and the interpretation and organization of experimental results. All AI-assisted mathematical derivations and proofs were manually verified and revised by the authors, and AI-assisted implementations were tested for correctness.

We did not use generative AI tools to develop the conceptual or methodological framework, formulate research hypotheses or mathematical claims, or design or provide feedback on the experiments. We also did not use generative AI tools to generate synthetic datasets, perform translation, clean or reformat datasets, or conduct qualitative or thematic data analysis.

Additionally, we used generative AI tools to assist with searching and summarizing related literature, manuscript drafting and organization, and improving readability and grammar.

We reviewed all AI-assisted work before incorporating it into this paper. Mathematical derivations and proofs were independently checked and revised by the authors, experimental implementations and scripts were tested for correctness, numerical results were verified against the corresponding experimental outputs, and references and manuscript claims were manually reviewed. We take responsibility for the final content of this work, including text, claims, code, analyses, and other artifacts produced with the aid of generative AI.

\subsection*{Reproducibility Statement}

We provide detailed experimental settings in Appendix~\ref{app:experimental-settings}, including data preprocessing, sparse representation learning, hierarchy construction, quality weighting, length-aware selection, and evaluation protocols. The appendix specifies the principal selection hyperparameters, benchmark suites, and aggregate metric definitions. Our experiments use publicly released baseline implementations, with downstream comparisons conducted under our reported training and evaluation settings. We will release our code and experimental configurations upon acceptance.

\subsection*{Ethics Statement}

This work is a fundamental algorithmic study of instruction data selection using publicly available datasets. It does not involve human-subject experiments or the collection of new personal data.

\bibliography{references}
\bibliographystyle{iclr2027_conference}

\appendix

\section{Appendix}

\subsection{Detailed Experimental Settings}
\label{app:experimental-settings}

\paragraph{Data and sparse representations.}
We use the Tulu~3 instruction pool containing $939{,}343$ conversations. After filtering examples whose serialized character length exceeds $22{,}500$, the eligible pool contains $929{,}647$ examples. Each complete conversation is encoded into a frozen $4096$-dimensional representation using Qwen3-8B. We adopt a three-resolution MSAE~\citep{bussmann2025learning} with nested widths of $1{,}024$, $4{,}096$, and $16{,}384$, using ReLU activations and global BatchTopK sparsification. We retain up to $64$ active features per example for subsequent hierarchy construction and selection.

\paragraph{Hierarchy and coverage.}
We construct the hierarchy using activation-support containment between adjacent resolutions. Each child concept retains its top $K_p=4$ candidate parents, with soft parent weights computed using temperature $T_p=0.1$. Ancestor propagation uses a depth-decay factor of $\rho=0.5$. For quality weighting, we use the released DEITA quality scorer~\citep{deita2024} and compute the expected score over its six output labels rather than taking the highest-scoring label. We normalize the resulting scores to $[0,1]$ and set the quality exponent to $\beta=0.5$. The concave coverage function uses $\gamma=0.85$.

\paragraph{Length-aware selection.}
We partition the eligible pool into $m=40$ quantile bins according to serialized conversation length. By default, we set $\tau=1$, allocating each bin a quota proportional to its mean length, with integer quotas normalized to the selection budget. Our primary experiments select $B=50{,}000$ examples. For stochastic greedy, we use $\varepsilon_{\mathrm{sg}}=0.1$, corresponding to at most $43$ sampled feasible candidates per iteration. The final selected index set $\mathcal I_B$ contains $50{,}000$ unique source indices.

\paragraph{SFT training.}
For the primary comparisons, we perform three-epoch full-parameter SFT with a 4K context window and sequence packing. All downstream SFT experiments are conducted on HC accelerator cards, whereas data selection and efficiency profiling are performed on a single NVIDIA B200 GPU with 32 CPU cores and 256\,GB of host memory.

\paragraph{Evaluation.}
We evaluate objective performance on ARC-Challenge, BBH, GSM8K, HumanEval pass@1, MMLU, and IFEval, reporting their unweighted mean as $\operatorname{Avg}_{\rm obj}$. Preference evaluation uses CompassJudger-1-32B on AlpacaEval-v2, MT-Bench, and WildBench. Following the comparison protocol, we compute
\begin{equation}
\operatorname{Avg}_{\rm sub}
=
\frac{\mathrm{AE}+10\,\mathrm{MT}+(\mathrm{Wild}+100)/2}{3},
\qquad
\operatorname{Avg}
=
\frac{\operatorname{Avg}_{\rm obj}+\operatorname{Avg}_{\rm sub}}{2}.
\end{equation}

\subsection{Additional Experimental Results}
\label{app:additional-models}

\paragraph{Different model initialization.}
Table~\ref{tab:qwen-instruct-main} extends the comparison to Qwen3-8B-Instruct. \method{} achieves the highest objective and subjective aggregates among the evaluated selectors, indicating that its selected subset remains effective when SFT starts from an instruction-tuned model rather than a Base initialization.

\begin{table*}[h]
	\centering
	\caption{Results on Qwen3-8B-Instruct under the 50K subset-selection setting. Full Tulu~3 is shown as a full-data reference and excluded from ranking. Best, second-best distinct, and third-best distinct 50K results are shown in bold, underline, and italics, respectively.}
	\label{tab:qwen-instruct-main}
	\begingroup
	\renewcommand{\arraystretch}{0.42}
	\resizebox{\textwidth}{!}{%
		\begin{tabular}{lllrrrrrrr|rrrrr}
			\toprule
			Base Model & Method & Data Size & ARC & BBH & GSM & HE & MMLU & IFEval & Avg$_{\rm obj}$ & AE & MT & Wild & Avg$_{\rm sub}$ & Avg \\
			\midrule
			\multirow{10}{*}{Qwen3-8B-Instruct}
			& Full Tulu~3 & 939K & 86.10 & 75.96 & 82.26 & 73.78 & 77.19 & 75.05 & 78.39 & 4.84 & 7.29 & -25.99 & 38.25 & 58.32 \\
			\cmidrule(l){2-15}
			& Uniform Random & 50K & \textit{89.49} & 79.65 & 84.91 & 76.83 & 76.44 & \textit{75.23} & 80.42 & 4.22 & 7.11 & -30.17 & 36.74 & 58.58 \\
			& ZIP & 50K & \underline{90.51} & 80.44 & \underline{87.34} & 36.59 & 76.53 & 72.83 & 74.04 & 3.23 & 7.03 & -42.51 & 34.09 & 54.07 \\
			& IFD & 50K & 87.12 & 80.53 & 83.02 & \textbf{82.32} & \underline{76.89} & 72.64 & 80.42 & \underline{9.19} & \textit{7.46} & \underline{-23.88} & \underline{40.62} & \underline{60.52} \\
			& \#InsTag & 50K & \textbf{91.19} & 79.10 & 83.47 & \textit{77.44} & \textit{76.72} & \textit{75.23} & 80.53 & 5.34 & 7.41 & -27.35 & 38.59 & 59.56 \\
			& DEITA & 50K & \underline{90.51} & 78.93 & \textbf{87.64} & 76.83 & 76.54 & 74.49 & \underline{80.82} & 5.47 & 7.15 & -32.88 & 36.84 & 58.83 \\
			& CaR & 50K & \underline{90.51} & 78.33 & 86.28 & 76.83 & \textbf{77.18} & 73.38 & 80.42 & 7.83 & \underline{7.47} & \textit{-25.95} & \textit{39.85} & 60.13 \\
			& QDIT & 50K & 89.15 & \textit{80.89} & 84.76 & \underline{78.05} & 76.64 & 74.49 & 80.66 & \textit{7.95} & \underline{7.47} & -26.51 & 39.80 & \textit{60.23} \\
			& MIG & 50K & 87.46 & \textbf{81.17} & \underline{87.34} & 75.61 & 75.75 & \textbf{76.71} & \textit{80.67} & \textit{7.95} & 7.40 & -28.24 & 39.28 & 59.98 \\
			\cmidrule(l){2-15}
			& \textbf{\method{}} & 50K & 89.15 & \underline{81.10} & \textit{86.50} & \underline{78.05} & 76.69 & \underline{75.97} & \textbf{81.24} & \textbf{9.94} & \textbf{7.75} & \textbf{-19.14} & \textbf{42.62} & \textbf{61.93} \\
			\bottomrule
		\end{tabular}%
	}
	\endgroup
\end{table*}

\begin{table*}[h]
	\centering
	\caption{Cross-family evaluation on Llama-3.1-8B and Mistral-7B-v0.3 under our reproduced 4K-context SFT protocol. All selection methods use 50K examples unless marked Full. Best, second-best distinct, and third-best distinct results among the evaluated 50K selectors are shown in bold, underline, and italics, respectively. Full Tulu~3 is a full-data reference and is excluded from ranking.}
	\label{tab:crossfamily-main}
	\begingroup
	\renewcommand{\arraystretch}{0.42}
	\resizebox{\textwidth}{!}{%
		\begin{tabular}{lllrrrrrrr|rrrrr}
			\toprule
			Base Model & Method & Data Size & ARC & BBH & GSM & HE & MMLU & IFEval & Avg$_{\rm obj}$ & AE & MT & Wild & Avg$_{\rm sub}$ & Avg \\
			\midrule
			\multirow{10}{*}{Llama-3.1-8B}
			& Full Tulu~3 & 939K & 80.68 & 69.27 & 76.65 & 62.20 & 66.14 & 67.65 & 70.43 & 6.58 & 6.75 & -27.70 & 36.74 & 53.59 \\
			\cmidrule(l){2-15}
			& Uniform Random & 50K & 77.97 & \underline{67.55} & 64.75 & \textbf{57.32} & 65.09 & \textit{54.71} & \textbf{64.56} & 7.08 & 6.88 & -35.42 & 36.06 & 50.31 \\
			& ZIP & 50K & 79.66 & 66.89 & 65.05 & 44.51 & \underline{65.78} & \underline{56.01} & 62.98 & 7.83 & 6.76 & -45.96 & 34.15 & 48.57 \\
			& IFD & 50K & \textit{80.00} & 65.58 & \textit{67.32} & \textit{52.44} & 65.32 & 49.72 & 63.40 & 12.17 & 6.85 & \underline{-25.99} & \textit{39.22} & 51.31 \\
			& \#InsTag & 50K & \textbf{83.39} & 67.41 & 65.43 & 47.56 & \textbf{66.09} & \textbf{56.19} & \underline{64.35} & 7.95 & 6.47 & -32.98 & 35.39 & 49.87 \\
			& DEITA & 50K & \textit{80.00} & 66.55 & \underline{67.55} & 46.95 & 65.05 & \underline{56.01} & 63.69 & 9.69 & 6.91 & -37.14 & 36.74 & 50.21 \\
			& CaR & 50K & \underline{82.03} & \textbf{68.25} & 66.11 & 51.83 & \textit{65.70} & 51.76 & \textit{64.28} & 11.43 & \underline{6.97} & -28.79 & 38.91 & \underline{51.60} \\
			& QDIT & 50K & 78.98 & 67.39 & 66.34 & \textit{52.44} & 65.38 & 51.94 & 63.74 & \textbf{14.04} & 6.81 & \textit{-28.03} & \underline{39.38} & \textit{51.56} \\
			& MIG & 50K & 69.15 & \textit{67.47} & \textbf{68.54} & \underline{53.05} & 64.71 & 52.87 & 62.63 & \textit{13.29} & \textit{6.95} & -31.94 & 38.94 & 50.79 \\
			\cmidrule(l){2-15}
			& \textbf{\method{}} & 50K & 72.88 & 65.75 & 65.35 & \underline{53.05} & 65.12 & \underline{56.01} & 63.03 & \underline{13.79} & \textbf{7.09} & \textbf{-25.59} & \textbf{40.63} & \textbf{51.83} \\
			\midrule
			\multirow{10}{*}{Mistral-7B-v0.3}
			& Full Tulu~3 & 939K & 35.25 & 58.72 & 72.55 & 62.80 & 61.55 & 73.01 & 60.65 & 9.07 & 6.97 & -22.54 & 39.17 & 49.91 \\
			\cmidrule(l){2-15}
			& Uniform Random & 50K & 75.59 & \textit{59.46} & 60.50 & \underline{46.34} & 60.60 & 64.33 & \underline{61.14} & 5.84 & \textit{6.99} & -28.78 & 37.12 & 49.13 \\
			& ZIP & 50K & \underline{78.98} & \underline{59.78} & 51.78 & 34.15 & 60.98 & 64.51 & 58.36 & 5.96 & 6.61 & -39.83 & 34.05 & 46.21 \\
			& IFD & 50K & 75.59 & 49.57 & 53.98 & 42.68 & 62.03 & 57.30 & 56.86 & \textit{14.16} & 6.72 & -22.88 & \textit{39.97} & 48.42 \\
			& \#InsTag & 50K & 76.27 & \textbf{59.81} & \underline{61.49} & 40.24 & 62.01 & \textit{66.91} & \textit{61.12} & 5.59 & \textbf{7.15} & -27.09 & 37.85 & 49.49 \\
			& DEITA & 50K & \textit{76.61} & 59.07 & \underline{61.49} & 38.41 & 61.38 & 62.85 & 59.97 & 6.58 & \textit{6.99} & -32.18 & 36.80 & 48.38 \\
			& CaR & 50K & \textit{76.61} & 59.06 & 59.51 & \textbf{48.17} & \underline{62.50} & 58.23 & 60.68 & 11.93 & 6.85 & -24.92 & 39.32 & 50.00 \\
			& QDIT & 50K & \textbf{80.00} & 59.02 & 60.80 & 40.24 & \textbf{62.91} & 61.55 & 60.75 & \underline{14.29} & 6.55 & \underline{-20.43} & 39.86 & \textit{50.31} \\
			& MIG & 50K & 72.20 & 58.48 & \textit{61.11} & \underline{46.34} & 61.39 & \textbf{68.95} & \textbf{61.41} & 13.17 & \underline{7.08} & \textit{-21.92} & \underline{41.00} & \underline{51.21} \\
			\cmidrule(l){2-15}
			& \textbf{\method{}} & 50K & 75.25 & 50.70 & \textbf{62.62} & \textit{45.73} & \textit{62.04} & \underline{67.65} & 60.66 & \textbf{15.53} & 6.96 & \textbf{-18.91} & \textbf{41.89} & \textbf{51.28} \\
			\bottomrule
		\end{tabular}%
	}
	\endgroup
\end{table*}

\paragraph{Cross-family evaluation.}
Table~\ref{tab:crossfamily-main} extends our evaluation to Llama-3.1-8B and Mistral-7B-v0.3. On Llama-3.1-8B, \method{} achieves the highest Final Avg among the evaluated 50K selectors, outperforming MIG by 1.04 points, with a more pronounced improvement in subjective evaluation. On Mistral-7B-v0.3, \method{} and MIG achieve comparable Final Avg, with \method{} achieving a slight 0.07-point advantage. These results indicate that the selected subset remains effective beyond the Qwen3 family, although the magnitude of the improvement varies across architectures.

Note that here we use the publicly released implementations of the competing methods and evaluate them under our unified training and evaluation protocol. All methods are retrained on our HC hardware with sequence packing to ensure consistent comparisons. The resulting scores can be a bit different from those reported in the original MIG paper, potentially due to differences in hardware, packing, or other implementation details. We therefore focus on comparisons within our experimental setting rather than directly comparing absolute scores with the published results.

{
\paragraph{Repeated-run evaluation.} We conduct 3-seed experiments to assess the robustness of H-MSAE across model scales and architectures, shown in Table. The additional results are reported as mean $\pm$ standard deviation over three independent training seeds. H-MSAE achieves positive mean Final Avg improvements over MIG on Qwen3-1.7B/4B/8B and Mistral-7B-v0.3, with gains of +1.05/+1.56/+0.62/+0.87 points, respectively.

\begin{table*}[!h]
\centering
{
\caption{{Three-seed downstream SFT comparison using 50K selected
examples. Values are mean $\pm$ standard deviation over three
training seeds. $\Delta$ denotes H-MSAE minus MIG in Final Avg,
computed from unrounded means.}}
\label{tab:hmsae_three_seed}
\small
\begin{tabular}{llcccc}
\toprule
Model & Method & Avg$_{\mathrm{obj}}\uparrow$ &
Avg$_{\mathrm{sub}}\uparrow$ & Final Avg$\uparrow$ &
$\Delta$ Final$\uparrow$ \\
\midrule
Qwen3-1.7B-Base & H-MSAE & \textbf{63.01 $\pm$ 0.36} &
\textbf{29.56 $\pm$ 0.45} & \textbf{46.29 $\pm$ 0.37} & -- \\
& MIG & 62.13 $\pm$ 0.32 & 28.34 $\pm$ 0.27 & 45.24 $\pm$ 0.11 & +1.05 \\
\midrule
Qwen3-4B-Base & H-MSAE & \textbf{77.62 $\pm$ 0.20} &
\textbf{40.11 $\pm$ 0.08} & \textbf{58.86 $\pm$ 0.13} & -- \\
& MIG & 76.67 $\pm$ 0.20 & 37.94 $\pm$ 0.54 & 57.30 $\pm$ 0.24 & +1.56 \\
\midrule
Qwen3-8B-Base & H-MSAE & 80.65 $\pm$ 0.28 &
\textbf{42.85 $\pm$ 0.16} & \textbf{61.75 $\pm$ 0.19} & -- \\
& MIG & \textbf{80.84 $\pm$ 0.09} & 41.42 $\pm$ 0.42 & 61.13 $\pm$ 0.18 & +0.62 \\
\midrule
Mistral-7B-v0.3 & H-MSAE & \textbf{61.27 $\pm$ 0.61} &
\textbf{42.88 $\pm$ 0.69} & \textbf{52.07 $\pm$ 0.65} & -- \\
& MIG & 60.97 $\pm$ 0.25 & 41.42 $\pm$ 0.51 & 51.20 $\pm$ 0.27 & +0.87 \\
\bottomrule
\end{tabular}
}
\end{table*}
}

{
\paragraph{Contribution of the hierarchical selection.}
We conduct experimental comparisons to justify the effect of hierarchical selection. The results are shown in Table \ref{tab:hmsae_hierarchy_ablation}.
Flat multi-resolution coverage improves over fine-only
coverage by +1.17/+0.93 Final Avg points on Qwen3-4B/8B.
H-MSAE further improves over flat multi-resolution by
+0.15/+0.23 points and exceeds shuffled-parent by
+0.39/+0.37 points. Thus, much of the gain comes from
multi-resolution coverage, while the inferred parent
relations contribute a smaller additional improvement.
The shuffled-parent control preserves the multi-resolution
representations and ancestor propagation, but randomizes
parent connections, supporting the relevance of the
inferred cross-resolution structure.
}

\begin{table*}[!h]
\centering
{
\caption{{Disentangling multi-resolution coverage and inferred
hierarchical relations. All runs select 50K examples and use
three-epoch SFT with three training seeds; values are mean
$\pm$ standard deviation. $\Delta$ Final denotes
H-MSAE minus the corresponding ablation and is computed
from unrounded means.}}
\label{tab:hmsae_hierarchy_ablation}
\small
\setlength{\tabcolsep}{4pt}
\begin{tabular}{llcccc}
\toprule
Model & Selector & Avg$_{\mathrm{obj}}\uparrow$ &
Avg$_{\mathrm{sub}}\uparrow$ & Final Avg$\uparrow$ &
$\Delta$ Final$\uparrow$ \\
\midrule
Qwen3-4B-Base & H-MSAE & \textbf{77.62 $\pm$ 0.20} &
\textbf{40.11 $\pm$ 0.08} & \textbf{58.86 $\pm$ 0.13} & -- \\
& Fine-only & 76.91 $\pm$ 0.47 & 38.16 $\pm$ 0.60 & 57.54 $\pm$ 0.18 & +1.32 \\
& Flat multi-resolution & 77.58 $\pm$ 0.25 & 39.85 $\pm$ 0.16 & 58.71 $\pm$ 0.20 & +0.15 \\
& Shuffled-parent & 77.29 $\pm$ 0.19 & 39.66 $\pm$ 0.40 & 58.48 $\pm$ 0.12 & +0.39 \\
\midrule
Qwen3-8B-Base & H-MSAE & \textbf{80.65 $\pm$ 0.28} &
\textbf{42.85 $\pm$ 0.16} & \textbf{61.75 $\pm$ 0.19} & -- \\
& Fine-only & 79.72 $\pm$ 0.24 & 41.48 $\pm$ 0.40 & 60.60 $\pm$ 0.31 & +1.16 \\
& Flat multi-resolution & 80.20 $\pm$ 0.11 & 42.85 $\pm$ 0.19 & 61.52 $\pm$ 0.09 & +0.23 \\
& Shuffled-parent & 80.11 $\pm$ 0.19 & 42.66 $\pm$ 0.33 & 61.39 $\pm$ 0.21 & +0.37 \\
\bottomrule
\end{tabular}
}
\end{table*}

{
\paragraph{Controlling length allocation.}
We compare H-MSAE with random selection and quality-only
selection using identical length-quantile bins and per-bin
integer quotas. Under this control, H-MSAE achieves a
+0.72-point higher mean Final Avg than quality-only and
+2.22 points over random selection. The quality-only gain
includes +1.21 objective points. These comparisons isolate
the length-allocation policy, although the selected subsets
need not contain identical numbers of supervised tokens.}

\begin{table}[!h]
\centering
{
\caption{{Length-quota-matched selection on Qwen3-4B-Base.
All methods use 50K selected examples, the same length bins
and integer quotas, and three-epoch SFT. Reported values are
mean $\pm$ standard deviation over three seeds.
$\Delta$ Final is H-MSAE minus baseline.}}
\label{tab:hmsae_length_controls}
\small
\setlength{\tabcolsep}{4pt}
\begin{tabular}{lcccc}
\toprule
Method & Avg$_{\mathrm{obj}}\uparrow$ &
Avg$_{\mathrm{sub}}\uparrow$ & Final Avg$\uparrow$ &
$\Delta$ Final$\uparrow$ \\
\midrule
H-MSAE & \textbf{77.62 $\pm$ 0.20} &
\textbf{40.11 $\pm$ 0.08} & \textbf{58.86 $\pm$ 0.13} & -- \\
Quality-only + quotas & 76.41 $\pm$ 0.28 & 39.86 $\pm$ 0.66 & 58.14 $\pm$ 0.43 & +0.72 \\
Random + quotas & 75.06 $\pm$ 1.19 & 38.22 $\pm$ 0.17 & 56.64 $\pm$ 0.66 & +2.22 \\
\bottomrule
\end{tabular}
}
\end{table}

{

\paragraph{Controlling supervised-token exposure.}
Under a matched budget of 123.73M supervised tokens
(the supervised token amount of H-MSAE in Table 1),
H-MSAE achieves 58.86 $\pm$ 0.13 Final Avg, compared
with 57.79 $\pm$ 0.32 for MIG, a +1.07-point mean gain.
This control shows that the aggregate difference is not
explained solely by higher supervised-token exposure.
Because optimizer-step counts and learning-rate trajectories
can differ, token matching does not independently control
every aspect of training compute.
}

\begin{table}[!h]
\centering

\caption{{Qwen3-4B-Base comparison under a matched
123.73M supervised-token budget. Values are mean
$\pm$ standard deviation over three training seeds.
$\Delta$ denotes H-MSAE minus MIG in Final Avg.}}
\label{tab:hmsae_token_controls}
\small
\setlength{\tabcolsep}{4pt}
\begin{tabular}{lcccc}
\toprule
Method & Avg$_{\mathrm{obj}}\uparrow$ &
Avg$_{\mathrm{sub}}\uparrow$ &
Final Avg$\uparrow$ &
$\Delta$ Final$\uparrow$ \\
\midrule
H-MSAE
& \textbf{77.62 $\pm$ 0.20}
& \textbf{40.11 $\pm$ 0.08}
& \textbf{58.86 $\pm$ 0.13}
& -- \\
MIG
& 77.00 $\pm$ 0.59
& 38.58 $\pm$ 0.19
& 57.79 $\pm$ 0.32
& +1.07 \\
\bottomrule
\end{tabular}
\end{table}

{
\paragraph{Comparison with SAE-based selection.}
We reproduce GreedSelect and SimScale from the selection
procedure in \citet{saediversity2025}, including length ordering
and iterative concept-set updates, and evaluate both
on the same 50K selection and SFT protocol.
H-MSAE outperforms GreedSelect/SimScale by +2.26/+2.73
Final Avg points on Qwen3-1.7B-Base and +3.12/+2.41
on Qwen3-4B-Base. This provides a direct empirical
comparison with SAE-based concept selection.
}

\begin{table*}[!h]
\centering
{
\caption{{Comparison with the SAE-based selection baselines
of Yang et al. (2025). All methods select 50K examples and
follow a three-epoch SFT protocol; values are mean $\pm$
standard deviation over three seeds. $\Delta$ Final is
H-MSAE minus the corresponding baseline.}}
\label{tab:hmsae_yang_comparison}
\small
\setlength{\tabcolsep}{4pt}
\begin{tabular}{llcccc}
\toprule
Model & Method & Avg$_{\mathrm{obj}}\uparrow$ &
Avg$_{\mathrm{sub}}\uparrow$ & Final Avg$\uparrow$ &
$\Delta$ Final$\uparrow$ \\
\midrule
Qwen3-1.7B-Base & H-MSAE & \textbf{63.01 $\pm$ 0.36} &
\textbf{29.56 $\pm$ 0.45} & \textbf{46.29 $\pm$ 0.37} & -- \\
& Yang GreedSelect & 61.31 $\pm$ 0.34 & 26.75 $\pm$ 0.85 & 44.03 $\pm$ 0.58 & +2.26 \\
& Yang SimScale & 59.85 $\pm$ 0.09 & 27.28 $\pm$ 0.41 & 43.56 $\pm$ 0.20 & +2.73 \\
\midrule
Qwen3-4B-Base & H-MSAE & \textbf{77.62 $\pm$ 0.20} &
\textbf{40.11 $\pm$ 0.08} & \textbf{58.86 $\pm$ 0.13} & -- \\
& Yang GreedSelect & 74.70 $\pm$ 0.57 & 36.77 $\pm$ 0.37 & 55.74 $\pm$ 0.39 & +3.12 \\
& Yang SimScale & 74.75 $\pm$ 0.35 & 38.14 $\pm$ 0.45 & 56.45 $\pm$ 0.07 & +2.41 \\
\bottomrule
\end{tabular}
}
\end{table*}

\paragraph{Additional data pools.}
We further evaluate \method{} against MIG on OpenHermes and Xsota to examine its effectiveness beyond the primary Tulu~3 pool. We use $\tau=2$ for this setting. As shown in Table~\ref{tab:additional-pools}, \method{} outperforms MIG on both OpenHermes and Xsota using Qwen3-4B.

\begin{table*}[h]
	\centering
	\small
	\setlength{\tabcolsep}{8pt}
	\caption{Additional evaluation on OpenHermes and Xsota. Final Avg is compared with MIG under the same model and data pool.
		 $\Delta$ is the difference relative to MIG.}
	\label{tab:additional-pools}
	\begin{tabular}{lllrrr}
		\toprule
		Model & Data pool & Configuration & \method{} & MIG & $\Delta$ \\
		\midrule
		Qwen3-4B-Base & OpenHermes & BF16 & 53.57 & 50.37 & +3.19 \\
		Qwen3-4B-Base & Xsota & BF16 & 53.93 & 52.20 & +1.74 \\
		\bottomrule
	\end{tabular}
\end{table*}

\begin{figure*}[h]
    \centering
    \includegraphics[width=\textwidth]{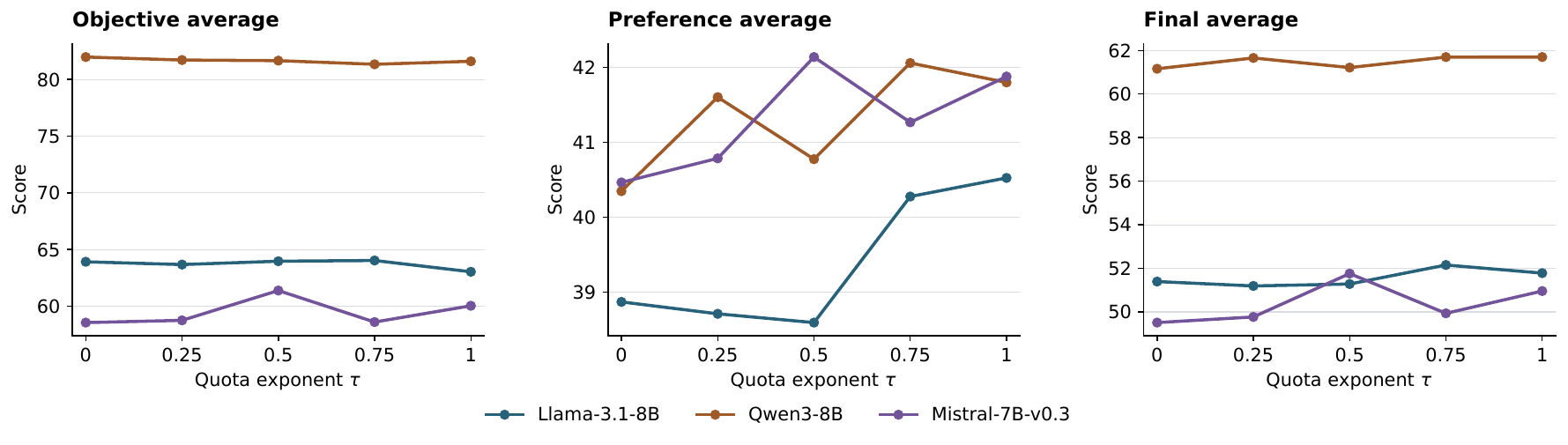}
    \caption{Sensitivity to the length-quota exponent $\tau$ across three model families. The panels report objective, subjective, and Final Avg scores for selection configurations with $\tau\in\{0,0.25,0.5,0.75,1\}$.}
    \label{fig:tau-sensitivity}
\end{figure*}

\paragraph{Sensitivity to length-stratification.}
We investigate the sensitivity of \method{} to the length-quota exponent $\tau$. As shown in Figure~\ref{fig:tau-sensitivity}, objective performance remains relatively stable across different values of $\tau$, while subjective performance exhibits greater variation and tends to stabilize around $\tau=0.75$. Despite these differences, Final Avg is comparatively insensitive to $\tau$, showing no consistent improvement with stronger length reweighting. These results suggest that \method{} is relatively robust to the choice of $\tau$ in terms of overall performance, although length allocation has a more noticeable effect on subjective evaluation.

\subsection{Proof of the Quality--Diversity Decomposition}
\label{app:quality-diversity-proof}

This section presents the modeling details and proof of Theorem~\ref{thm:quality-diversity}, using the notation introduced in Section~\ref{sec:subset-utility}.

\paragraph{Modeling details.}
We model the covariance of each stochastic training gradient as
\begin{equation}
\operatorname{Cov}(g_i)
=
(1-q_i)\Sigma,
\qquad
\Sigma\succeq0,
\label{eq:appendix-quality-covariance}
\end{equation}
where $q_i\in[0,1]$ represents latent example quality and $\Sigma$ is a shared covariance matrix. Thus, higher-quality examples have lower gradient uncertainty. We assume uncorrelated gradient fluctuations across distinct examples:
\begin{equation}
\operatorname{Cov}(g_i,g_j)
=
\mathbb E\!\left[
(g_i-\bar g_i)(g_j-\bar g_j)^\top
\right]
=
0,
\qquad i\neq j.
\label{eq:appendix-cross-covariance}
\end{equation}

To isolate the effects of quality and directional diversity, we consider a homogeneous local regime:
\begin{equation}
g_{\mathcal T}^{\top}\bar g_i
=
\mu,
\qquad
\bar g_i^\top H_{\mathcal T}\bar g_i
=
\chi>0,
\qquad \forall i.
\label{eq:appendix-homogeneous-regime}
\end{equation}
These conditions fix the first-order target alignment and self-curvature across candidate examples. We further define
\begin{equation}
\nu
=
\operatorname{tr}(H_{\mathcal T}\Sigma)
\ge0.
\label{eq:appendix-nu}
\end{equation}

Finally, we assume that $R_{\mathcal T}$ has a locally Lipschitz-continuous Hessian and that the stochastic subset-average gradient has a finite third moment. These conditions ensure that the expected remainder in the second-order Taylor expansion of $R_{\mathcal T}(\theta_0+\Delta_S)$ is $o(\eta^2)$ as $\eta\to0$, for each fixed subset $S$.

\paragraph{An equivalent form of directional diversity.}
Because $\|v_i\|_2=1$ and
$D(S)=\operatorname{tr}(\widehat{\operatorname{Cov}}_{i\in S}[v_i])$
uses the $1/(B-1)$ sample-covariance normalization,
\begin{equation}
D(S)
=
1-
\frac{2}{B(B-1)}
\sum_{\substack{i,j\in S\\i<j}}
v_i^\top v_j.
\label{eq:appendix-diversity-identity}
\end{equation}
Equivalently,
\begin{equation}
2\sum_{\substack{i,j\in S\\i<j}}
v_i^\top v_j
=
B(B-1)\bigl(1-D(S)\bigr).
\label{eq:appendix-pairwise-sum}
\end{equation}
This identity is used only to simplify the cross-example terms below.

\paragraph{Proof of Theorem~\ref{thm:quality-diversity}.}
Applying a second-order Taylor expansion of the target risk around $\theta_0$ to the local subset update, we obtain
\begin{equation}
U_{\mathcal T}(S)
=
-g_{\mathcal T}^{\top}\Delta_S
-\frac{1}{2}\Delta_S^\top H_{\mathcal T}\Delta_S
+
o(\eta^2).
\label{eq:appendix-utility-taylor}
\end{equation}

For the first-order term,
\begin{align}
-\mathbb E[g_{\mathcal T}^{\top}\Delta_S]
&=
\frac{\eta}{B}
\sum_{i\in S}
g_{\mathcal T}^{\top}\bar g_i 
=
\eta\mu.
\label{eq:appendix-first-order}
\end{align}

For the second-order term,
\begin{align}
\mathbb E[\Delta_S^\top H_{\mathcal T}\Delta_S]
&=
\frac{\eta^2}{B^2}
\left[
\sum_{i\in S}
\mathbb E[g_i^\top H_{\mathcal T}g_i]
+
2\sum_{\substack{i,j\in S\\i<j}}
\mathbb E[g_i^\top H_{\mathcal T}g_j]
\right].
\label{eq:appendix-second-order}
\end{align}

For each diagonal term,
\begin{align}
\mathbb E[g_i^\top H_{\mathcal T}g_i]
&=
\bar g_i^\top H_{\mathcal T}\bar g_i
+
\operatorname{tr}\!\left(
H_{\mathcal T}\operatorname{Cov}(g_i)
\right) 
=
\chi+(1-q_i)\nu.
\end{align}
Therefore,
\begin{equation}
\sum_{i\in S}
\mathbb E[g_i^\top H_{\mathcal T}g_i]
=
B\chi+B\nu\bigl(1-Q(S)\bigr).
\label{eq:appendix-diagonal-sum}
\end{equation}

For $i\neq j$, Eq.~\eqref{eq:appendix-cross-covariance} gives
\begin{equation}
\mathbb E[g_i^\top H_{\mathcal T}g_j]
=
\bar g_i^\top H_{\mathcal T}\bar g_j.
\end{equation}
From the definition
$v_i=H_{\mathcal T}^{1/2}\bar g_i/
\|H_{\mathcal T}^{1/2}\bar g_i\|_2$
and the homogeneous self-curvature assumption,
\begin{align}
v_i^\top v_j
&=
\frac{
\bar g_i^\top H_{\mathcal T}\bar g_j
}{
\sqrt{\bar g_i^\top H_{\mathcal T}\bar g_i}
\sqrt{\bar g_j^\top H_{\mathcal T}\bar g_j}
} 
=
\frac{
\bar g_i^\top H_{\mathcal T}\bar g_j
}{\chi}.
\end{align}
Hence,
\begin{equation}
\bar g_i^\top H_{\mathcal T}\bar g_j
=
\chi v_i^\top v_j.
\label{eq:appendix-cross-direction}
\end{equation}
Using Eq.~\eqref{eq:appendix-pairwise-sum},
\begin{equation}
2\sum_{\substack{i,j\in S\\i<j}}
\mathbb E[g_i^\top H_{\mathcal T}g_j]
=
\chi B(B-1)\bigl(1-D(S)\bigr).
\label{eq:appendix-cross-sum}
\end{equation}

Combining Eqs.~\eqref{eq:appendix-diagonal-sum} and
\eqref{eq:appendix-cross-sum} yields
\begin{align}
\mathbb E[\Delta_S^\top H_{\mathcal T}\Delta_S]
&=
\eta^2\chi
+
\frac{\eta^2\nu}{B}
-
\frac{\eta^2\nu}{B}Q(S)
-
\frac{\eta^2\chi(B-1)}{B}D(S).
\end{align}

Substituting into Eq.~\eqref{eq:appendix-utility-taylor},
\begin{align}
\mathbb E[U_{\mathcal T}(S)]
&=
\eta\mu
-\frac{\eta^2\chi}{2}
-\frac{\eta^2\nu}{2B}
+\frac{\eta^2\nu}{2B}Q(S)
+\frac{\eta^2\chi(B-1)}{2B}D(S)
+o(\eta^2) \\
&=
C_B+\lambda_QQ(S)+\lambda_DD(S)+o(\eta^2),
\end{align}
where
\begin{equation}
C_B
=
\eta\mu-\frac{\eta^2\chi}{2}-\frac{\eta^2\nu}{2B},
\qquad
\lambda_Q
=
\frac{\eta^2\nu}{2B}
\ge0,
\qquad
\lambda_D
=
\frac{\eta^2\chi(B-1)}{2B}
>0.
\end{equation}
This proves Theorem~\ref{thm:quality-diversity}.
\hfill$\square$

\end{document}